\documentclass[10pt,twocolumn,letterpaper]{article}

\usepackage[pagenumbers]{cvpr}

\definecolor{TeseoColor}{rgb}{0.15, 0.68, 0.38}
\definecolor{NafisehColor}{rgb}{1.0, 0.0, 0.5}

\definecolor{codegreen}{rgb}{0,0.6,0}
\definecolor{codegray}{rgb}{0.5,0.5,0.5}
\definecolor{codepurple}{rgb}{0.58,0,0.82}
\definecolor{backcolour}{rgb}{0.95,0.95,0.95}

\usepackage{listings}

\lstdefinestyle{mystyle}{
    backgroundcolor=\color{backcolour},   
    commentstyle=\color{codegreen},
    keywordstyle=\color{magenta},
    numberstyle=\tiny\color{codegray},
    stringstyle=\color{codepurple},
    basicstyle=\ttfamily\tiny,
    breakatwhitespace=false,         
    breaklines=true,                 
    captionpos=b,                    
    keepspaces=true,                 
    numbers=none,                    
    numbersep=0pt,    
    xleftmargin=0pt,
    xrightmargin=0pt,
    showspaces=false,                
    showstringspaces=false,
    showtabs=false,                  
    tabsize=1,
    breakatwhitespace=true
}

\definecolor{cvprblue}{rgb}{0.21,0.49,0.74}
\usepackage[pagebackref,breaklinks,colorlinks,allcolors=cvprblue]{hyperref}

\title{LLM-Guided Material Inference for 3D Point Clouds}

\author{Nafiseh Izadyar\\
University of Victoria\\
\and
Teseo Schneider\\
University of Victoria\\
}

\begin{document}
\maketitle

\begin{abstract}
Most existing 3D shape datasets and models focus solely on geometry, overlooking the material properties that determine how objects appear. We introduce a two-stage large language model (LLM) based method for inferring material composition directly from 3D point clouds with coarse segmentations. Our key insight is to decouple reasoning about what an object is from what it is made of. In the first stage, an LLM predicts the object's semantic; in the second stage, it assigns plausible materials to each geometric segment, conditioned on the inferred semantics. Both stages operate in a zero-shot manner, without task-specific training. Because existing datasets lack reliable material annotations, we evaluate our method using an \emph{LLM-as-a-Judge} implemented in DeepEval \cite{Ip_deepeval_2025}. Across 1,000 shapes from Fusion \cite{willis2022joinable}/ABS \cite{izadyar2025better} and ShapeNet \cite{chang2015shapenet}, our method achieves high semantic and material plausibility. These results demonstrate that language models can serve as general-purpose priors for bridging geometric reasoning and material understanding in 3D data.
\end{abstract}

\section{Introduction}
\label{sec:intro}

The past decade has witnessed the creation of numerous large-scale 3D shape datasets, such as ShapeNet, ABC, Fusion, ABS, and PointNet \cite{chang2015shapenet, koch2019abc, willis2022joinable, izadyar2025better, qi2017pointnet}.
These datasets have driven significant progress in data-driven shape analysis; most notably in 3D reconstruction \cite{huang2023neural}, segmentation \cite{fu2023bpnet, kolodiazhnyi2024oneformer3d}, and shape generation \cite{yang2024pscad, liu2025hola, Li_2025_CVPR}. Despite these advances, a crucial dimension of real 3D shape remains overlooked: \emph{material appearance}. Most publicly available shape datasets lack annotations on how the object looks or what it is made of, limiting their ability to support applications that depend on material and appearance understanding. For instance, it is fundamental for photorealistic rendering necessary for robotics reinforcement learning, where reflectance and texture determine visual realism.

While several recent works have explored visual material recognition from RGB images \cite{fang2024make}, estimating material composition directly from 3D geometry shapes remains an open challenge. The difficulty stems from the ambiguity of 3D data without colour or material information, the diversity of materials, and the limited availability of annotated 3D datasets. This makes it hard to apply traditional supervised methods that rely on dense, per-part material labels. Fortunately, many practical 3D shape representations already include coarse structural segmentation (e.g., CAD models contain shells, meshes contain face groups). This segmenting information often corresponds to material boundaries and can be exploited for material estimation.

\begin{figure}
\centering\footnotesize
\includegraphics[width=\linewidth]{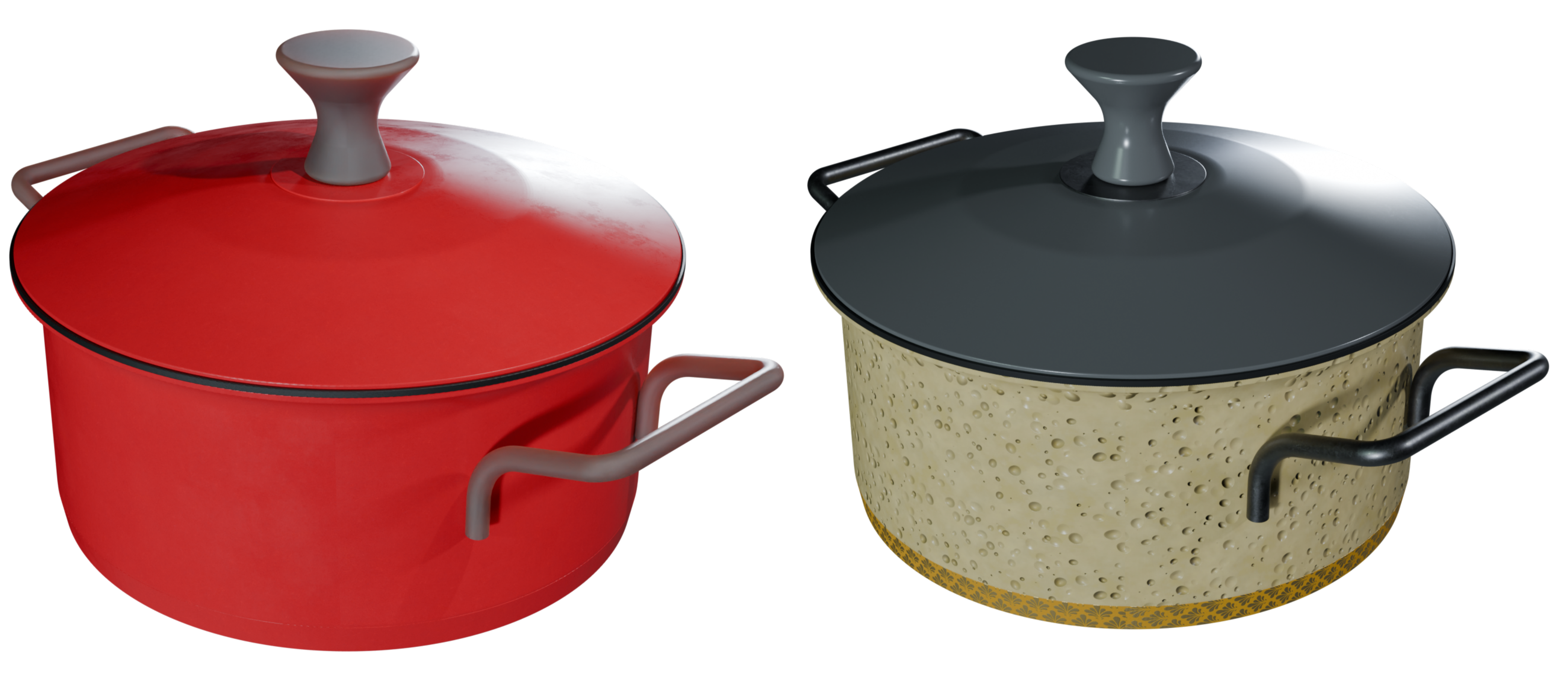}
\caption{A pot-shaped object misclassified without semantic reasoning. Directly inferring materials from geometry assigns incorrect materials (i.e., foam, fabric), left. Our pull pipeline first identifies the object as a pot and then assigns plausible materials such as metal, plastic, and rubber.}
\label{fig:example}
\end{figure}

\begin{figure*}
\centering\footnotesize
\includegraphics[width=\linewidth]{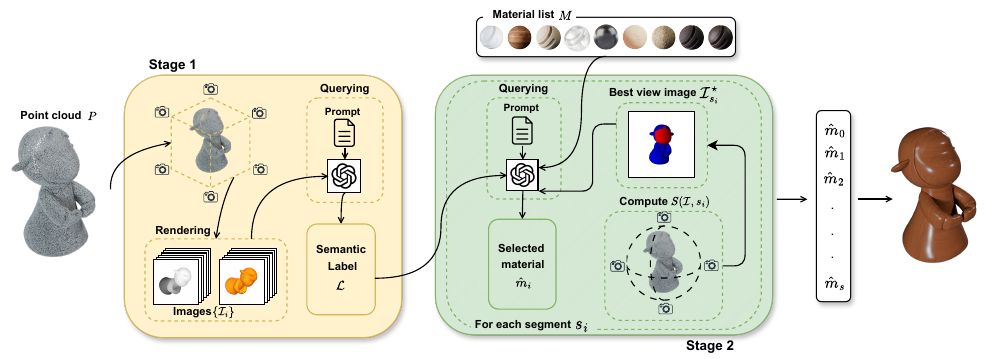}
\caption{Overview of our method:
Stage~1 uses an LLM to infer high-level semantic information about the object from multi-view depth and raster renderings;
Stage~2 combines the semantic label with the input coarse segmentation to predict a material for each part.}
\label{fig:pipeline}
\end{figure*}

 In this paper, we propose a language-model-driven approach to infer material composition directly from this coarse 3D representation. Our key insight is that large language models (LLMs) can recognize shapes and can reason about plausible material assignments based on object semantics and part relationships (Fig.~\ref{fig:example}). Given a point cloud segmented into a small number of parts, our method predicts plausible materials for each part. This enables robust, interpretable material inference even in the absence of explicit training labels and supervision.

We validate our method (Section~\ref{sec:eval}) on over 1,000 models from the Fusion/ABS and ShapeNet datasets, using the LLM as the judge \cite{zheng2023judging}. Across diverse object categories and origins, our approach achieves high accuracy in recovering the correct material composition. We also provide qualitative visualizations that show our predictions align closely with human perception of material plausibility.
In summary, our contributions are twofold:
\begin{enumerate}
    \item We introduce the first LLM-based framework for inferring material composition directly from segmented 3D geometry.
    \item We demonstrate that our approach achieves high material classification accuracy, establishing a new baseline for material-aware 3D understanding.
\end{enumerate}

\section{Related Work}

\paragraph{3D Shape Understanding and Representation Learning.}
Large 3D datasets such as ShapeNet~\cite{chang2015shapenet}, ModelNet~\cite{wu20153d}, and the ABC dataset~\cite{koch2019abc} have driven rapid progress in geometry-centric learning for classification, segmentation, and reconstruction.  
Point-cloud and mesh encoders including PointNet~\cite{qi2017pointnet}, DGCNN~\cite{wang2019dynamic}, and subsequent transformer-based architectures~\cite{guo2021pct} have become standard for geometric reasoning tasks.  
While these approaches learn powerful shape descriptors, they capture only geometric structure omitting appearance, material, and physical context that are crucial for realistic perception and interaction.  
Recent works such as PointNeXt~\cite{qian2022pointnext} and SparseConvNet~\cite{graham20183d} further improve geometric fidelity but remain limited to purely spatial features.

\paragraph{Material and Appearance Estimation.}
Estimating material properties has a long history in computer vision.  
Classical inverse-rendering methods recover reflectance and BRDF parameters from RGB images~\cite{rematas2016deep, li2018learning, deschaintre2020physically, guo2021materialgan, li2018learning1}, while material recognition from photographs uses convolutional or multimodal networks~\cite{bell2015material, hu2018directional}. For 3D data, however, most methods assume textured meshes or images, not bare geometry. Some works attempt to infer material categories from surface shape cues~\cite{schwartz2019recognizing, milan2020material}, but rely on curated datasets with aligned appearance supervision. Our work addresses a different setup: infer materials directly from untextured, coarsely segmented point clouds.

\paragraph{Language-Driven and Multimodal 3D Understanding.}
Recent advances in vision-language modeling extend material and semantic reasoning beyond pixels. Models such as CLIP~\cite{radford2021clip} and BLIP-2~\cite{li2023blip2} enable zero-shot recognition by aligning visual and textual representations. Their 3D counterparts ULIP~\cite{xue2023ulip}, CLIP-Forge~\cite{sanghi2022clipforge}, and OpenShape~\cite{liu2023openshape} project 3D features into shared language spaces for zero-shot shape retrieval and captioning. These models reason about category-level semantics but not physical materials. More recently, large language models (LLMs) have been used as general reasoning engines for 3D understanding (e.g., LLM-Grounder~\cite{hong2023llmgrounder} and Chat-3D~\cite{chen2024chat3d}) where text-based reasoning augments 3D perception. Our approach follows this trend, employing an LLM not as a language–vision aligner but as a semantic prior to infer plausible materials from geometric context.

\paragraph{Part-Aware 3D Reasoning.}
Part-level understanding underpins both semantic segmentation and physical reasoning. Large-scale part-annotated datasets such as PartNet~\cite{mo2019partnet} enable hierarchical decomposition of shapes into functional substructures, inspiring recent neural architectures for part segmentation~\cite{yu2021pointr, hu2022scanobjectnn}. Although part-aware learning captures object organization, prior methods treat parts as purely geometric entities. We extend this line by associating parts with material plausibility, bridging shape decomposition and appearance inference.

\paragraph{Relation to CAD and Structural Reconstruction.}
CAD-oriented methods~\cite{li2019spfn, sharma2020parsenet, cao2020cadgraph} focus on analytic surface reconstruction and design-intent recovery, which share our emphasis on geometric structure but differ in goal: reconstructing parametric surfaces rather than interpreting material semantics. Our method complements these by introducing language-based reasoning into geometry-only data, enabling semantic and material inference without any color, texture, or supervision.

\section{Method}\label{sec:method}

\begin{figure}
    \centering
    \includegraphics[width=.33\linewidth]{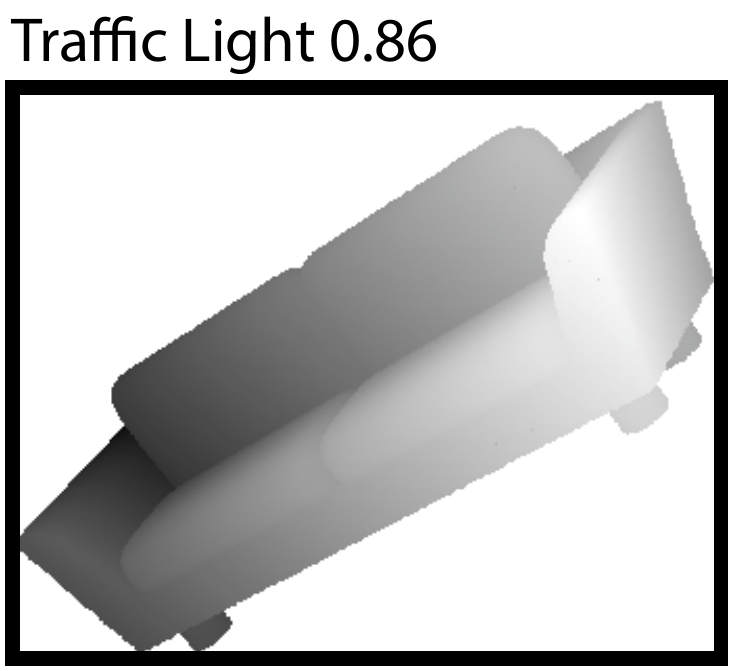}\hfill
    \includegraphics[width=.33\linewidth]{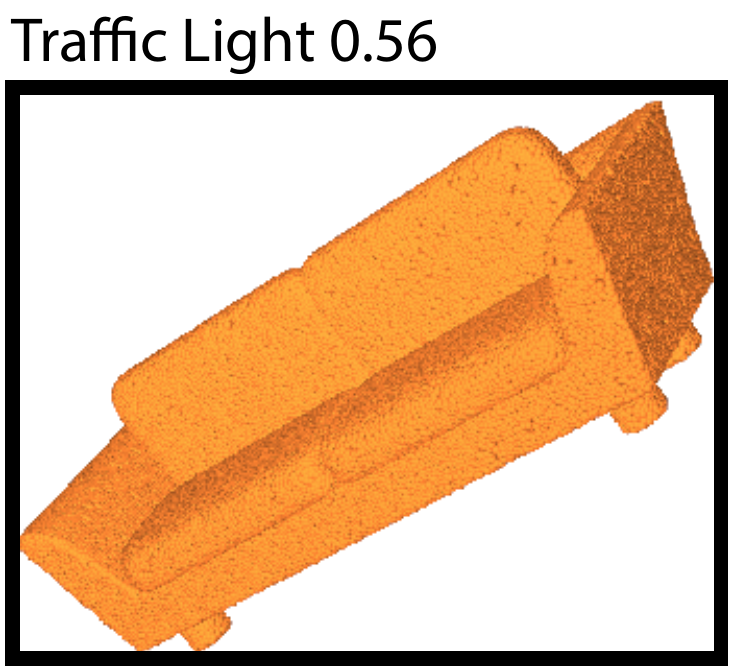}\hfill
    \includegraphics[width=.33\linewidth]{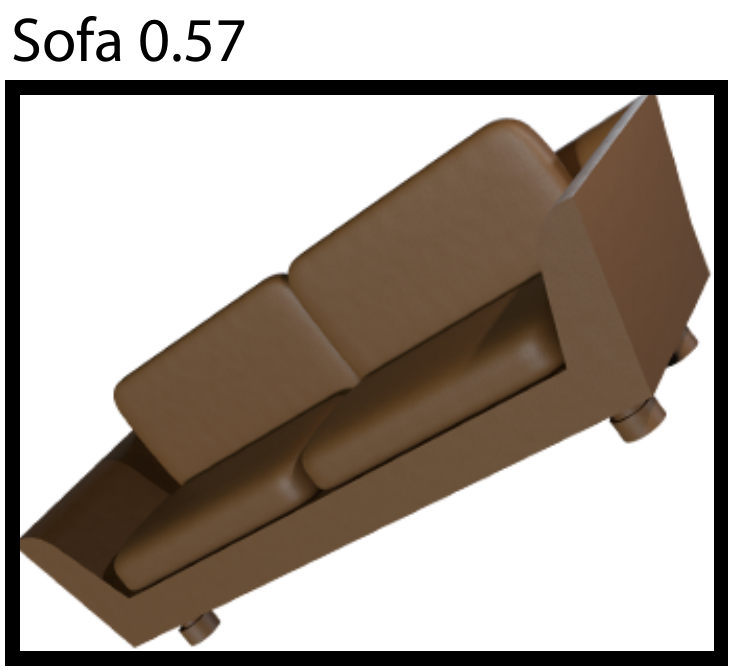}\hfill
    \caption{Existing state-of-the-art \cite{Zhao_2024_CVPR} object classification methods fail on our raster or depth-only inputs, as they rely heavily on texture, color, or contextual cues; the couch is detected as a traffic light (left and middle). By assigning the correct material (right) the object is correctly identified.}
    \label{fig:sotafail}
\end{figure}

\paragraph{Overview.}
Large language models implicitly encode rich semantic and physical priors about the world, learned from large-scale text training. They ``know'' that a saucepan handle is often plastic or wood, or that a car wheel rim is metal while the tire is rubber. We leverage this linguistic commonsense, combined with geometric reasoning, to infer materials directly from object semantics and structure. 
However, without semantic cues, this task is highly ambiguous (Section~\ref{sec:ablation}). For example, when attempting to assign materials to a pot shape without semantics, the system incorrectly predicts foam and fabric (Fig.~\ref{fig:example}, left). Inferring materials solely from geometry is therefore unreliable. 

To overcome this limitation, we explicitly separate reasoning about \emph{what} the object is from reasoning about \emph{what it is made of}, which we implement as a two-stage pipeline (Fig.~\ref{fig:pipeline}): 
\begin{enumerate}
\item Stage 1 \emph{Semantic extraction} (Section~\ref{sec:step1}) detects object semantic information.
\item Stage 2 \emph{Material inference} (Section~\ref{sec:step2}) combines the predicted semantics with the coarse geometric segmentation to assign materials to each part.
\end{enumerate}
This separation allows our system to correctly identify the object as a pot and assign plausible materials: metal, plastic, and rubber for the gasket (Fig.~\ref{fig:example}, right).

\paragraph{Input and output.}
Our input is a point cloud $P=\{p_i\}$ $i=1,\dots, n$ with $n$ points with a coarse segmentation into $k$ segments $S=\{s_i\}$, $i=1,\dots,k$, where every segment $s_i$ contains a subset of $P$ and $s_i \cap s_j = \emptyset$ if $i\neq j$. 
Such segmentations can originate from CAD hierarchies (shells), meshes (face groups), or automatic segmentation algorithms \cite{fu2023bpnet, kolodiazhnyi2024oneformer3d}. Additionally, we assume access to a discrete set of $m$ candidate materials $ M = \{m_i\}$, $i=1, \dots, m$.
The goal is to predict one material label $\hat{m}_k\in M$ for each segment $s_i$ representing the estimated material composition of the object.

\begin{figure}
  \centering\footnotesize
    \includegraphics[width=.33\linewidth]{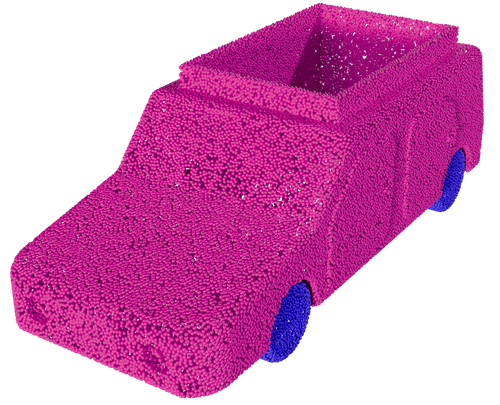}\hfill
    \includegraphics[width=.33\linewidth]{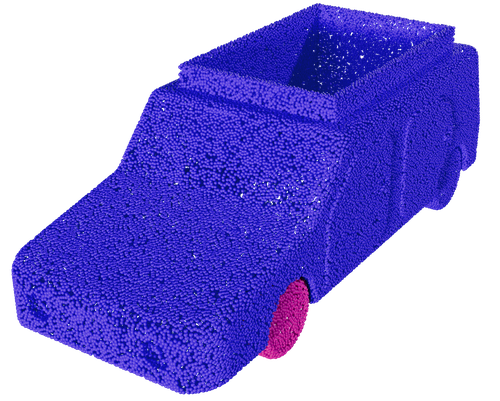}\hfill
    \includegraphics[width=.33\linewidth]{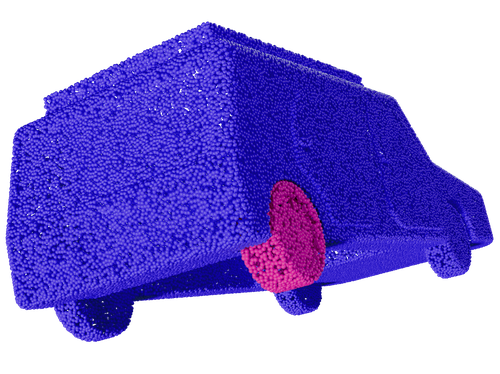}
  \caption{Example of best view for several highlighted segments in pink.}
  \label{fig:segexample}
\end{figure}

\begin{figure}
\centering\footnotesize
\includegraphics[width=\linewidth]{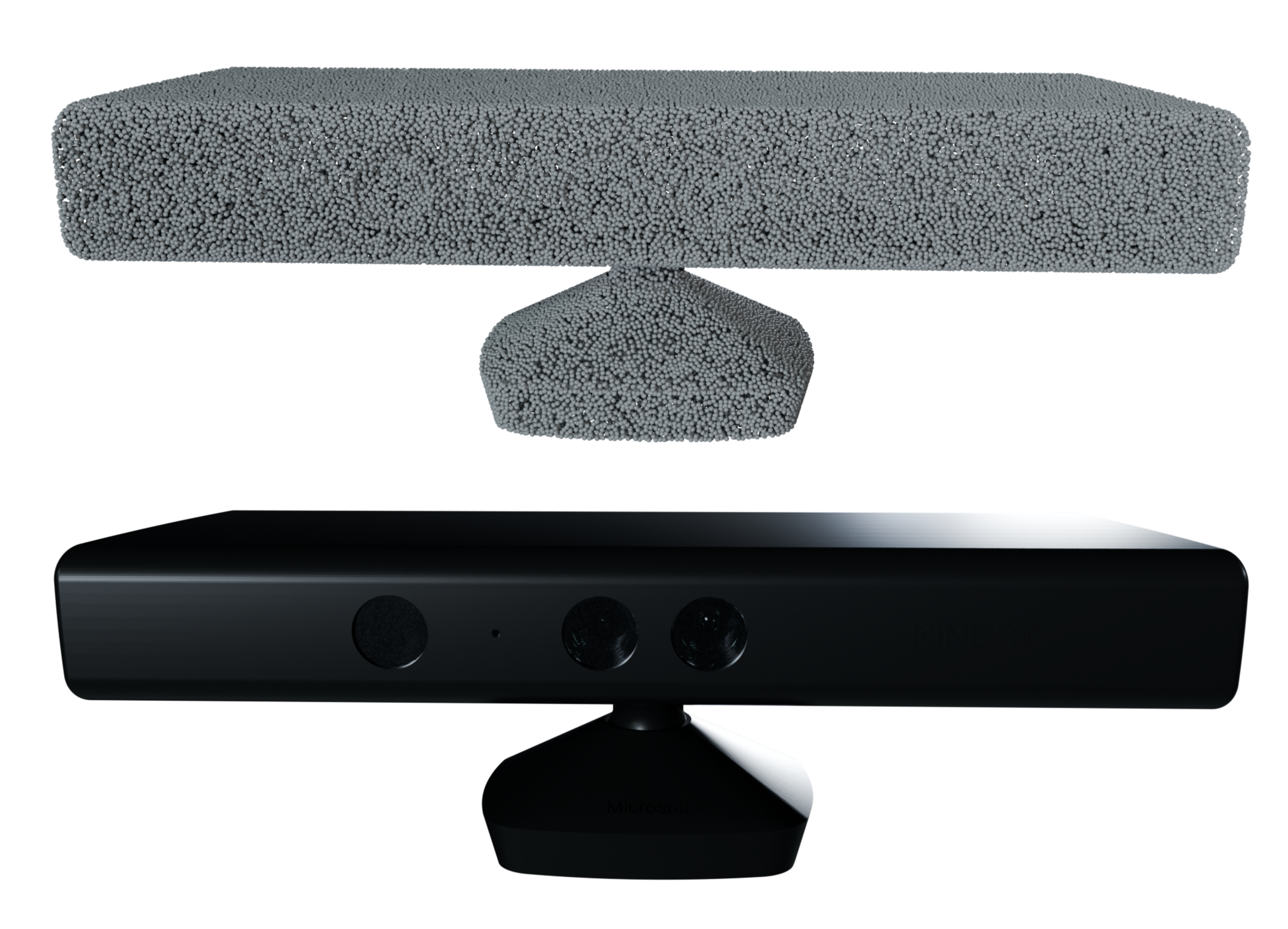}
\caption{
Example of an object (top) from the Fusion dataset that clearly represents a Microsoft Kinect, but the material assignments are incorrect. In the data, the whole object is metal and plastic. Our method correctly assigns glass (for the lenses) and plastic (for the body) (bottom).}
\label{fig:bad-fusion}
\end{figure}

\begin{figure*}
  \centering\footnotesize
  \includegraphics[width=\linewidth]{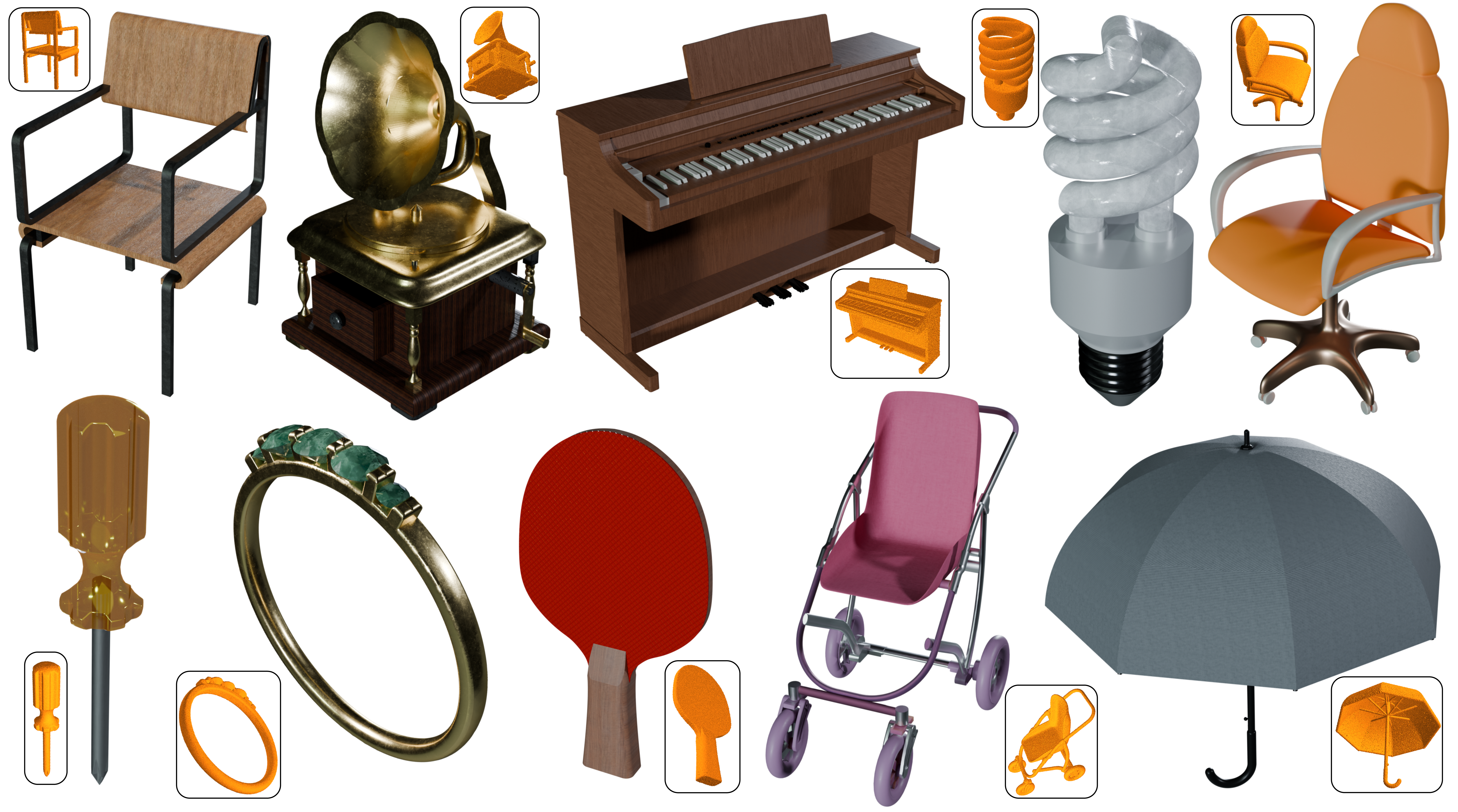}

  \caption{Qualitative results of our method across various datasets, in orange we show the input point cloud.}
  \label{fig:results}
\end{figure*}

\subsection{Stage 1: Object semantics}\label{sec:step1}
The goal of the first stage is to extract semantic information about the object; that is, determining what the model represents.
Since we only have geometric information (represented by the point cloud), existing 3D classification methods cannot be applied, as they typically rely on texture/color, global context, or orientation, which are absent in our data (Fig.~\ref{fig:sotafail}). 

To overcome this limitation, we develop an LLM-based approach for semantic extraction. We render the input point cloud into $\ell=8$ depth and raster images $\mathcal{I}_i$ each corresponding to one corner of a cube of the cube aligned with the principal directions determined via principal component analysis. 
We found that using eight views strikes a good balance: they provide sufficient coverage without significant loss in performance (Section~\ref{sec:ablation}).

The resulting set of rendered depth and raster images $\{\mathcal{I}_i\}$ is passed to an LLM together with Prompt~\ref{semantic-prompt}, which outputs both a semantic label $\mathcal{L}$ and a confidence score. If the confidence falls below $\varepsilon_s$, the input is marked as a semantic failure and excluded from Stage 2.

\begin{lstlisting}[breakindent=0pt, caption={Prompt used in the semantic extraction step of our method.}, label={semantic-prompt}]
   
   You are an object detector. You will receive 16 renderings of the same CAD     model:
     - 8 depth maps (grayscale depth; no color/texture).
     - 8 point-cloud projections.
     
     Camera is placed at the 8 corners of a cube around the object. So from         each viewpoint you have two images: one depth and one rendered point           clouds.

    Task:
    - Output one candidate class label for the object that best describes it.
    - Output a confidence score as an integer 0-100.
    - If uncertain, you may output "unknown" (confidence <= 30).

    Labeling Guidelines:
    - Try to be rotation- and scale-invariant. Do not infer                         "front/back/top/bottom" from pose.
    - Base your judgment only on geometry. Ignore color entirely.
    - If the geometry is featureless, low-detail, or highly ambiguous, reduce       confidence accordingly; use "unknown" when appropriate.
    - Keep it to 1-2 words.

    Confidence Calibration:
    - 80-100: Distinctive geometry, minimal ambiguity, very confident.
    - 55-79: Reasonable match with some uncertainty but still detectable.
    - 31-54: Vague cues; object likely but not certain, unsure.
    - <=30: Insufficient/ambiguous; consider "unknown".

    OUTPUT (STRICT):
    Return ONLY valid JSON matching this exact shape and length:
    [
        {{ "answer": "label", "confidence": 0 }}
    ]
\end{lstlisting}

\subsection{Stage 2: Material estimation}\label{sec:step2}
The goal of the second stage is to assign a material label $\hat{m}_i \in M$ to each segment $s_i$. For each segment, we render the point cloud, highlighting $s_i$ (Fig.~\ref{fig:segexample}). Because not all viewpoints clearly expose the target segment and provide the same semantic information, we design a heuristic to select the best view.

For a rendered image $\mathcal{I}$ and segment $s_i$, we compute three quantities: the Shannon entropy of the unique color distribution in Lab color space
\[
E_s(\mathcal{I}, s_i) = \sum_{k=1}^{K} 
      \frac{n_k}{N}
      \log\Big(\frac{n_k}{N}\Big),
\]
with $n_k$ be the number of pixels assigned each unique color (histogram), and $N$ be the total number of non-white pixels. 
The depth score
\[
D_s(\mathcal{I}, s_i) = \text{std} \big(\mathcal{I}_d\big),
\]
where $\mathcal{I}_d$ is the depth rendering of $\mathcal{I}$.
Finally, the visibility score
\[
V_s(\mathcal{I}, s_i) = p_s/p_t,
\]
with $p_s$ the number of pixel in the segment and $p_t$ the total number of pixels excluding the background.
These are combined into a single image score
\[
S(\mathcal{I}, s_i) = \alpha E_s(\mathcal{I}, s_i)+\beta D_s+\gamma V_s(\mathcal{I}, s_i),
\]
where we choose the weights $\alpha,\beta,\gamma$ empirically (Section~\ref{sec:ablation}).
We sample $c$ random camera positions uniformly on a unit sphere, compute $S(\mathcal{I}, s_i)$ for each camera, and select the best-view image 
\[
\mathcal{I}^\star_{s_i} = \arg\max_\mathcal{I} S(\mathcal{I}, s_i).
\]
To avoid images where $s_i$ is not visible, we discard any view with $V_s(\mathcal{I}, s_i) < \varepsilon_v = 1e-5$.

Finally, we query the LLM with $\mathcal{I}^\star_{s_i}$, the object semantic label $\mathcal{L}$, and the candidate materials $M$ using Prompt~\ref{material-prompt}; the model returns the most plausible material $\hat{m}_i$ for $s_i$.

\begin{lstlisting}[breakindent=0pt, caption={Prompt used in the material estimation step of our method.}, label={material-prompt}]

    You are a materials and design engineering expert. You are given the CAD       object name, list of available materials, and one rendered image from the      CAD model.
        
    TASK:
    - Your need to assign the most appropriate material to the                     red-highlighted part of the given rendered image of the CAD model.
        
    RULES:
    1) Do not invent new materials; only select from the provided                  materials list.
    2) You need to choose a material that is commonsense for example power         cord cannot be metal.
    3) If multiple options fit, break ties in this order: mechanical               performance, environment/temperature/chemicals, manufacturability,              cost/availability, mass/finish.
    4) If VERY uncertain, you can select "unknown".
    5) Consider only the red-highlighted region in the image.
            
    INFORMATION:
    - Object name is: f{object_type}.
    - Available materials list (select one string exactly as written):               f{material_list}.
        
    OUTPUT (STRICT):
    Return ONLY this JSON with no extra text:
        [ 
            {{"material":"from materials list","confidence": 0 }}
        ]
                
\end{lstlisting}

\section{Evaluation}\label{sec:eval}

\begin{figure}
  \centering\footnotesize
   \includegraphics[width=.33\linewidth]{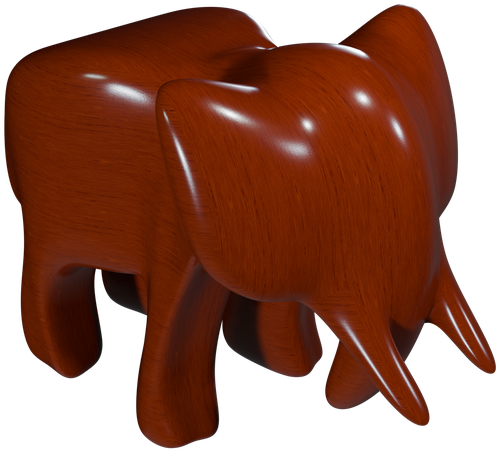} \hfill
    \includegraphics[width=.33\linewidth]{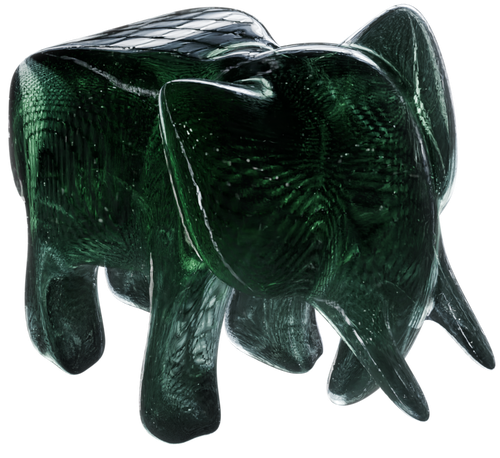}\hfill
    \includegraphics[width=.33\linewidth]{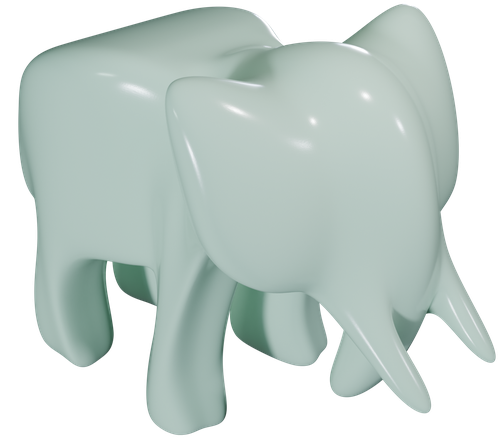}
  \caption{Same model with different candidate sets $M$. From left to right, using the default materials the elephant is wood, by removing wood from the list it is detected as glass. Finally, by removing both, our method assign porcelain.}
  \label{fig:diff-mat}
\end{figure}

\begin{figure*}
  \centering\footnotesize
    \includegraphics[width=\linewidth]{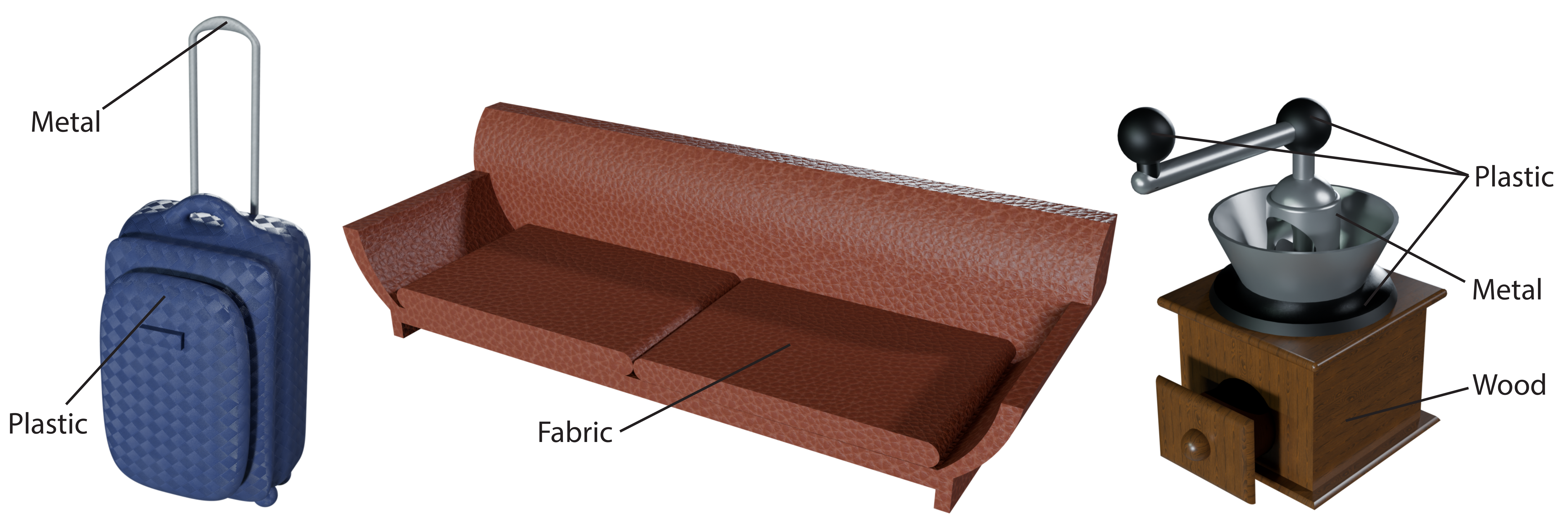}
  \caption{Asymmetric shape that cannot be correctly recognized with only four views.  With eight views, Stage~1 correctly identifies them.}
  \label{fig:asym}
\end{figure*}

Since no reliable ground-truth material annotations exist for our datasets, we adopt an \emph{LLM-as-a-Judge} evaluation framework~\cite{chiang2023llm_eval}, implemented using DeepEval ~\cite{Ip_deepeval_2025}.  Although the ABS/Fusion dataset includes material labels, we found many of them incorrect (Fig.~\ref{fig:bad-fusion}).
Therefore, we rely on two independent LLM-based judges. Each judge responds with a binary decision (\texttt{correct} or \texttt{incorrect}) to assess the quality of the two stages. The full text prompts used for both judges are provided in the Supplementary material. As a baseline, we used the single-stage variant that predicts materials directly from images without semantic reasoning; it performs significantly worse (Section~\ref{sec:ablation}).

\paragraph{Stage 1.}
The first judge evaluates the performance of the semantic extraction phase.
We provide the judge with all eight depth maps and eight rendered images used in Stage~1 together with the semantic label $\mathcal{L}$ predicted by our method, and we ask whether the label is correct. We define the \emph{model confidence} $M_c$ as the percentage of models that produce a label for $\varepsilon_s = 70\%$, and the \emph{semantic accuracy} $S_a$ as the percentage of models whose predicted label is judged correct.

\paragraph{Stage 2.}
The second judge evaluates the material inference phase and determines whether the assigned material is plausible. We provide the judge with the object semantic label $\mathcal{L}$, the list of candidate materials $M$, and for each segment $s_i$, its best-view rendering $\mathcal{I}^\star_{s_i}$ and assigned material $\hat{m}_i$. Note that when $\mathcal{L}$ is judged incorrectly by the first judge, we replace it with ``unknown.''
We compute two metrics: \emph{material accuracy} $M_a$, the percentage of models for which all segments are judged correct, and \emph{mean segment accuracy} $\sigma_a$, the average fraction of correctly judged segments per model.

\paragraph{Quantitative Results.}
We evaluate our method on $1,000$ 3D point clouds with at most 10 segments (for performance reason) drawn from the ABS and ShapeNet datasets, and we use $M$ as: Metal, Wood, Stone, Glass, Ceramic, Plastic, Rubber, Foam, Fabric and Paper
(Fig.~\ref{fig:results}). The choice of candidate materials $M$ is part of the input and directly affects the resulting assignments  (Fig.~\ref{fig:diff-mat}). In our experiments we use $\alpha=1$, $\beta=1$, and $\gamma=100$ to favor views where the segment is clearly visible (Section~\ref{sec:ablation}). Our method successfully predicts valid semantic labels for $\varepsilon_v = 70\%$ for 88.09\% of the models. Among these, the first-stage semantic accuracy, as judged by the LLM, is 72.52\% (Table~\ref{tab:stage1}). The accuracy of the second stage is similar: 81.43\% models have all segments correctly assigned, and on average 93.06\% of segments have plausible material (Table~\ref{tab:stage2}). These results demonstrate that our two-stage framework reliably infers both object semantics and material plausibility across diverse datasets, even in the absence of ground-truth annotations.

\begin{table}
    \centering
    \caption{Stage~1 performance across datasets.}
    \label{tab:stage1}
    \begin{tabular}{lcc}
        \toprule
        Dataset & $M_c$ (\%) & $S_a$ (\%) \\
        \midrule
        ABS \cite{izadyar2025better}/Fusion\cite{willis2022joinable} &  89.13 & 73.58 \\
        ShapeNet \cite{chang2015shapenet} & 87.06 & 71.46 \\
        \bottomrule
    \end{tabular}
\end{table}

\begin{table}
    \centering
    \caption{Stage~2 material accuracy across datasets.}
    \label{tab:stage2}
    \begin{tabular}{lcc}
        \toprule
        Dataset & $M_a$ (\%) & $\sigma_a$ (\%) \\
        \midrule
        ABS \cite{izadyar2025better}/Fusion\cite{willis2022joinable} & 83.74 & 95.61 \\
        ShapeNet \cite{chang2015shapenet} & 79.12 & 90.52 \\
        \bottomrule
    \end{tabular}
\end{table}

\paragraph{Runtime and Cost Analysis.}
We evaluate the computational cost of our full pipeline using the GPT-4.1 model. For each 3D point cloud, Stage~1 requires eight depth maps, eight raster images, and one LLM query, while Stage~2 performs one query per segment (five segments per model on average). On a standard workstation, rendering the 16 views takes 2.09s in average in total, and LLM inference takes around 11.60s. The second stage has similar timing, resulting in a total average runtime of 23.29s per model. The evaluation is cheap: around 5.13M tokens for both judges per 1,000 models. Table~\ref{tab:costs} summarizes these results, where $Q$ denotes the average number of queries per model, $t$ the mean runtime, and $\tau$ the number of input tokens per model.

\section{Ablation}\label{sec:ablation}

We conduct a series of ablation experiments to evaluate the impact of individual design choices in our method. All experiments use the same LLM configuration and judging procedure described in Section~\ref{sec:eval}. We run the ablation on 200 randomly selected models from our combined dataset.

\paragraph{Effect of Semantic Reasoning.}
We disable Stage~1 of our pipeline and attempt to estimate materials directly from images \emph{without} first extracting semantic information. With this change, the overall material accuracy $M_a$ drops to 7.35\% (from 80.43\% with the full two-stage pipeline). This confirms our central hypothesis: separating \emph{what} an object is from \emph{what it is made of} is crucial for accurate material inference.

\paragraph{Raster and Depth.}
In Stage~1, we provide both a rendered image and a depth map. With $\varepsilon_s = 70\%$ confidence threshold, using depth alone reduces the fraction of models identified from 88.10\% (passing both the depth map and rendered image) to 21.92\%.  This suggests that the LLM relies on both geometric and visual cues to identify objects accurately. Eliminating depth leads to roughly a 10\% drop in the success rate of the identification.

\paragraph{Number of Rendered Views.}
As described in Section~\ref{sec:method}, we render eight principal depth and raster views for semantic extraction.  
Reducing this number to four or increasing it to twelve alters the performance by at most 4.5\%.
Four views often suffice for symmetric shapes but fail for highly asymmetric ones (Fig.~\ref{fig:asym}).

\begin{figure}
  \centering\footnotesize
    \includegraphics[width=\linewidth]{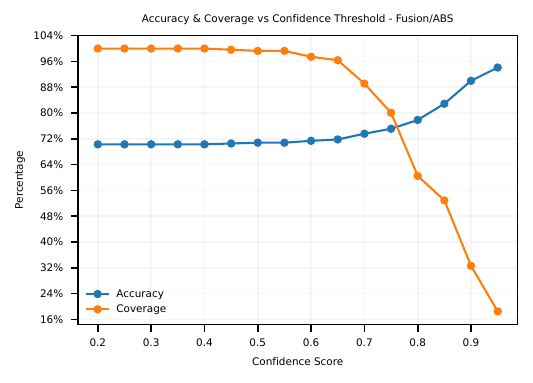}\hfill
    \includegraphics[width=\linewidth]{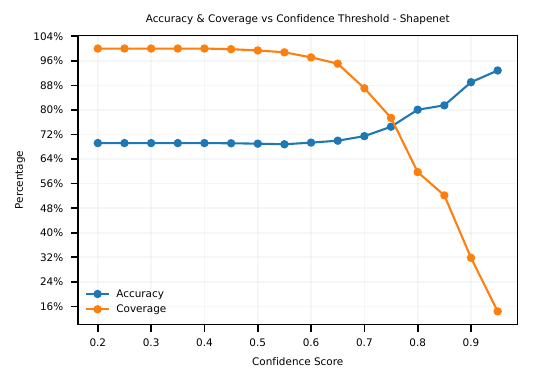}\hfill
  \caption{Performance of our method for varying confidence thresholds in Stage~1.}
  \label{fig:confidence}
\end{figure}

\begin{table}
    \centering
    \caption{Runtime and cost analysis of our method.}
    \label{tab:costs}
    \begin{tabular}{lccc}
        \toprule
        Stage & $Q$  & $t$ (s) & $\tau$ \\
        \midrule
        Stage 1 & 1 & 13.69 & 4409 \\
        Stage 2 & 5 & 9.6 & 544 \\
        \midrule
        \textbf{Total} & 6 & 23.29 & 4953 \\
        \bottomrule
    \end{tabular}
\end{table}

\paragraph{LLM Confidence Threshold.}
We exclude all semantic labels whose LLM confidence falls below $\varepsilon_s=70\%$ (Section~\ref{sec:method}). Fig.~\ref{fig:confidence} shows how accuracy varies with this threshold. Accepting low-confidence labels introduces incorrect semantics that propagate to material inference, while a stricter threshold reduces the model confidence ($M_c$ decreases to 52.5\%). A threshold of 70\% offers the best trade-off between accuracy and coverage.

\paragraph{Weights in $S(\mathcal{I}, s_i)$.}
We varied the relative weights in $S(\mathcal{I}, s_i)$ but did not observe significant differences (about 3\% change in $M_a$).
This suggests the heuristic combination of entropy, depth, and visibility is robust. However, if we include images with low visibility scores $V_s(\mathcal{I}, s_i)$ (i.e., set $\gamma = 0$) and allow views where segments might be occluded, $M_a$ drops to 74\%.

\paragraph{Number of Sampled Cameras.}
To obtain the optimal viewpoint $\mathcal{I}^\star_{s_i}$, we sample $c=10$ camera positions uniformly on the unit sphere. Increasing to $c=50$ alters $S(\mathcal{I}, s_i)$ only marginally (about $2\!-\!3\%$) and yields nearly identical viewpoints, with negligible effect on $M_a$.

\section{Conclusions}

\begin{figure}
  \centering\footnotesize
    \includegraphics[width=.45\linewidth]{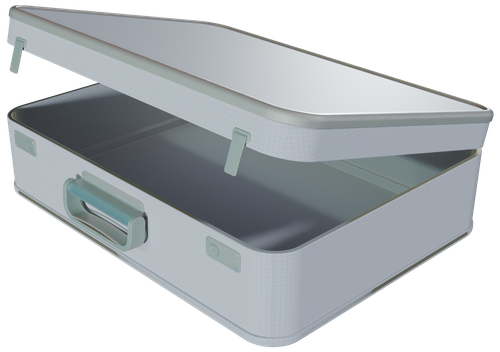}\hfill
    \includegraphics[width=.45\linewidth]{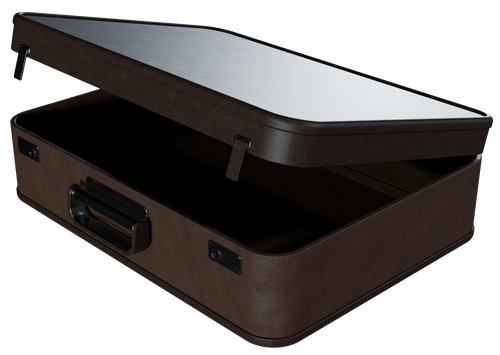}
  \caption{A briefcase incorrectly labeled as a newspaper (left), leading Stage~2 to assign paper materials. Manually correcting the label enables Stage 2 to correctly assign materials (right).}
  \label{fig:briefcase}
\end{figure}

We presented a two-stage LLM-driven method for material estimation from coarsely segmented 3D point clouds, demonstrating that combining geometric semantics with language-based reasoning enables accurate and interpretable predictions without supervision. 
Our experiments show that semantic understanding is essential for reliable material estimation, and that large language models can serve as effective priors for appearance reasoning. 
Beyond point-cloud material inference, the same framework could generalize to functional and structural understanding; for instance, reason about how parts move or interact. 
We note that Stage~1 is imperfect (Figure~\ref{fig:briefcase}, left), highlighting an interesting research avenue: accurately identifying shapes without access to color, scale, or contextual cues. 
Because this stage is independent of Stage~2, improving object recognition would directly enhance overall material accuracy (Fig.~\ref{fig:briefcase}, right).

While our approach demonstrates robust material reasoning without supervision, it also inherits several limitations. 
It relies on the common sense priors encoded in the underlying language model; consequently, incorrect or culturally biased associations (e.g., rare or ambiguous materials) may propagate to the predictions. Similarly, since our evaluation depends on LLM-based judges rather than human annotations, it may has similar biases or inconsistencies in edge cases. Finally, our method assumes the presence of a coarse but meaningful segmentation; applying it to raw, unsegmented point clouds would not work. We believe that creating a new method to create material-semantic annotations represents an exciting venue for future work.

We believe this work opens a new direction for language-guided 3D perception, where common sense knowledge complements geometric analysis to achieve richer, more human-like understanding of physical objects.

{}

\end{document}